\documentclass{article}
\usepackage{arxiv}
\usepackage[utf8]{inputenc} 
\usepackage[T1]{fontenc}    
\usepackage[hyperfootnotes=false]{hyperref}       
\usepackage{url}            
\usepackage{booktabs}       
\usepackage{amsfonts}       
\usepackage{nicefrac}       
\usepackage{microtype}      
\usepackage{lipsum}		
\usepackage{graphicx}
\usepackage{grffile}
\usepackage[numbers]{natbib}
\usepackage{doi}
\usepackage{amsmath}
\usepackage{amsthm}
\usepackage{subcaption}
\usepackage{footnote}
\makesavenoteenv{table*}
\makesavenoteenv{tabular}
\usepackage{tablefootnote}
\usepackage{mwe}
\usepackage{float}
\usepackage{ragged2e}
\floatstyle{plaintop}
\restylefloat{table}
\usepackage{makecell}

\usepackage{etoolbox}

\title{OmniFusion Technical Report}

\author{{\hspace{1mm}Elizaveta Goncharova} \\
	\And
	{\hspace{1mm}Anton Razzhigaev} \\
        \And
	{\hspace{1mm}Matvey Mikhalchuk} \\
        \And
	{\hspace{1mm}Maxim Kurkin} \\
        \And
	{\hspace{1mm}Irina Abdullaeva} \\
        \And
        {\hspace{1mm}Matvey Skripkin} \\
        \And
        {\hspace{1mm}Ivan Oseledets} \\
        \And
	{\hspace{1mm}Denis Dimitrov} \\
        \And
        {\hspace{1mm}Andrey Kuznetsov} \\
}

\renewcommand{\shorttitle}{OmniFusion Technical Report}

\begin{document}
\maketitle

\begin{abstract}

Last year, multimodal architectures served up a revolution in AI-based approaches and solutions, extending the capabilities of large language models (LLM). We propose an \textit{OmniFusion} model based on a pretrained LLM and adapters for visual modality. We evaluated and compared several architecture design principles for better text and visual data coupling: MLP and transformer adapters, various CLIP ViT-based encoders (SigLIP, InternVIT, etc.), and their fusing approach, image encoding method (whole image or tiles encoding) and two 7B LLMs (the proprietary one and open-source Mistral). Experiments on 8 visual-language benchmarks show the top score for the best OmniFusion setup in terms of different VQA tasks in comparison with open-source LLaVA-like solutions: VizWiz, Pope, MM-Vet, ScienceQA, MMBench, TextVQA, VQAv2, MMMU. We also propose a variety of situations, where OmniFusion provides highly-detailed answers in different domains: housekeeping, sightseeing, culture, medicine, handwritten and scanned equations recognition, etc. Mistral-based OmniFusion model is an open-source solution with weights, training and inference scripts available at \href{https://github.com/AIRI-Institute/OmniFusion} {https://github.com/AIRI-Institute/OmniFusion}.

\end{abstract}

\begin{figure}[h!]
    \centering
    \includegraphics[scale=.5]{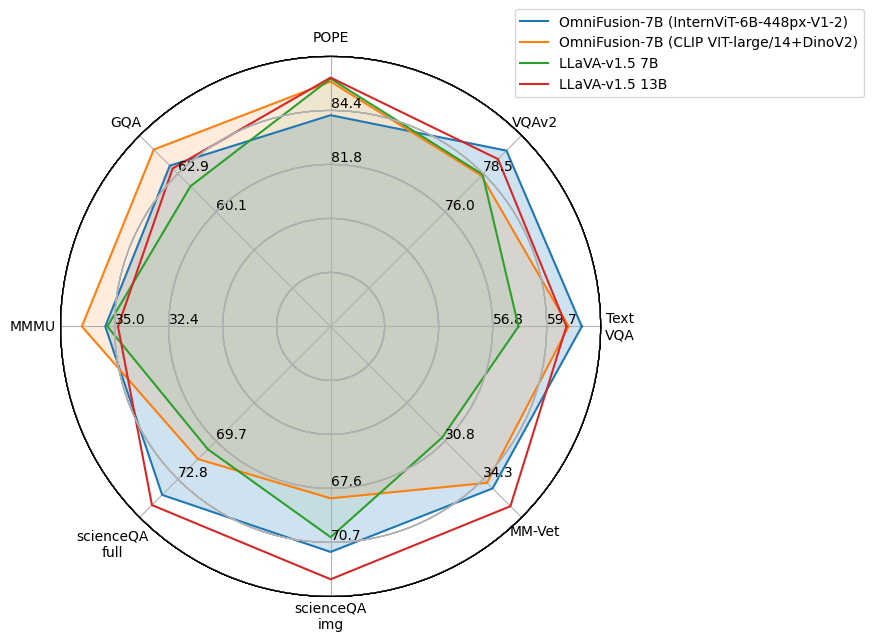}
    \caption{Comparison of OmniFusion performance on the benchmarks and generation examples.}
    \label{fig:radar-plot}
\end{figure}

\begin{figure}[h!]
    \centering
    \includegraphics[scale=.9]{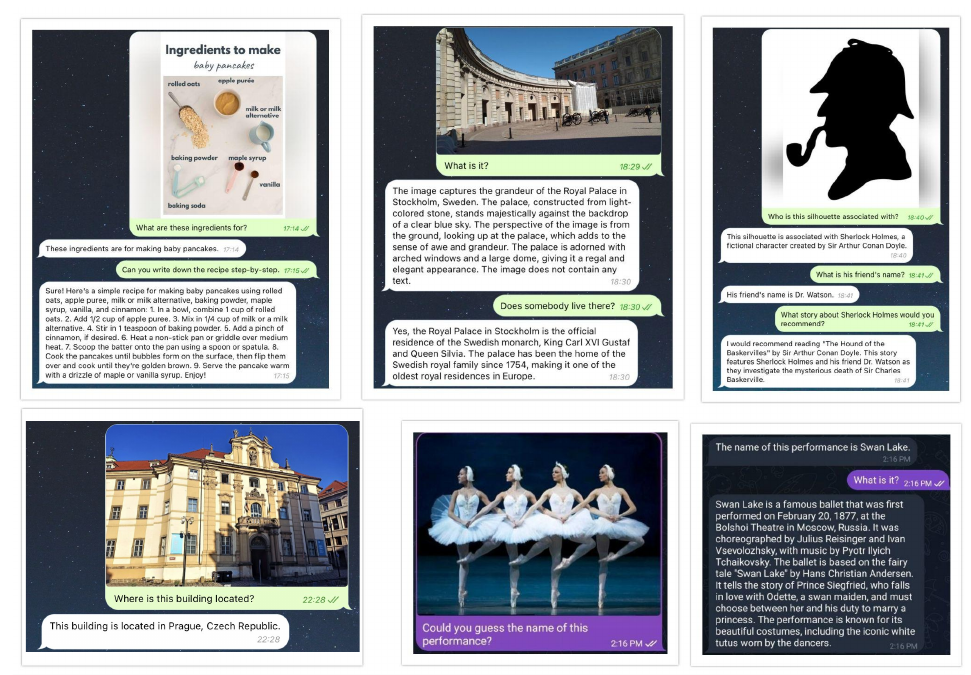}
    \caption{OmniFusion VQA examples.}
    \label{fig:radar-plot}
\end{figure}

\section{Introduction}

In recent years, multimodal architectures emerged as a powerful paradigm for enhancing artificial intelligence (AI) systems, enabling them to process and understand multiple types of data simultaneously \cite{Alayrac2022FlamingoAV, Gao2023LLaMAAdapterVP, Lin2023VideoLLaVALU}. The integration of different data modalities, such as text and images, has significantly improved the capabilities of large language models (LLMs) in various tasks, ranging from visual question answering (VQA) \cite{Liu2023LLaVAPlusLT} to complex decision-making processes \cite{Li2023LLaVAMedTA, gemini_vs_gpt}. However, the challenge of effectively coupling various data types remains a significant obstacle in the development of truly integrative AI models. Furthermore, such multimodal multitask architectures are interpreted as the first steps towards the development of the artificial general intelligence (AGI), expanding the number of challenges in world cognition.

This work introduces the \textit{OmniFusion} model, a novel multimodal architecture that leverages the strengths of pretrained LLMs and introduces specialized adapters for processing visual information. We evaluated multiple architectural designs to fuse text and visual data, such as MLP and transformer adapters. At the same time we concentrated on the comparison of various image encoders like CLIP-ViT \cite{radford2021learning} and SigLIP \cite{zhai2023sigmoid} and on visual encoders fusion techniques. The evaluated approaches gave us a broad scope of improving the visual context for better visual information extraction and retrieval.

One of the key innovations of OmniFusion is its flexible approach to image encoding, exploring both the whole image and the tiled image encoding strategies, which allows for a more nuanced understanding of visual content in relation to textual data \cite{liu2024llavanext}. This adaptability is critical in addressing the diverse requirements of visual-language benchmarks, where OmniFusion has demonstrated superior performance across a range of VQA tasks, outperforming existing open-source solutions.

Our extensive evaluations on eight visual-language benchmarks, including VizWiz \cite{gurari2018vizwiz}, POPE \cite{li2023evaluating}, MM-Vet \cite{yu2023mmvet}, ScienceQA \cite{lu2022learn}, MMBench \cite{liu2023mmbench}, TextVQA \cite{singh2019vqa}, VQAv2 \cite{vqav2}, and MMMU \cite{yue2023mmmu}, confirm the effectiveness of the OmniFusion model. The model not only achieves competitive performance with the leading approaches in various VQA tasks but also excels in providing detailed answers across multiple domains, such as housekeeping, sightseeing, culture, and medicine, showcasing its versatility and broad applicability.

\section{OmniFusion}

\subsection{Model Architecture}

\begin{figure*}
    \centering
    \includegraphics[width=\linewidth]{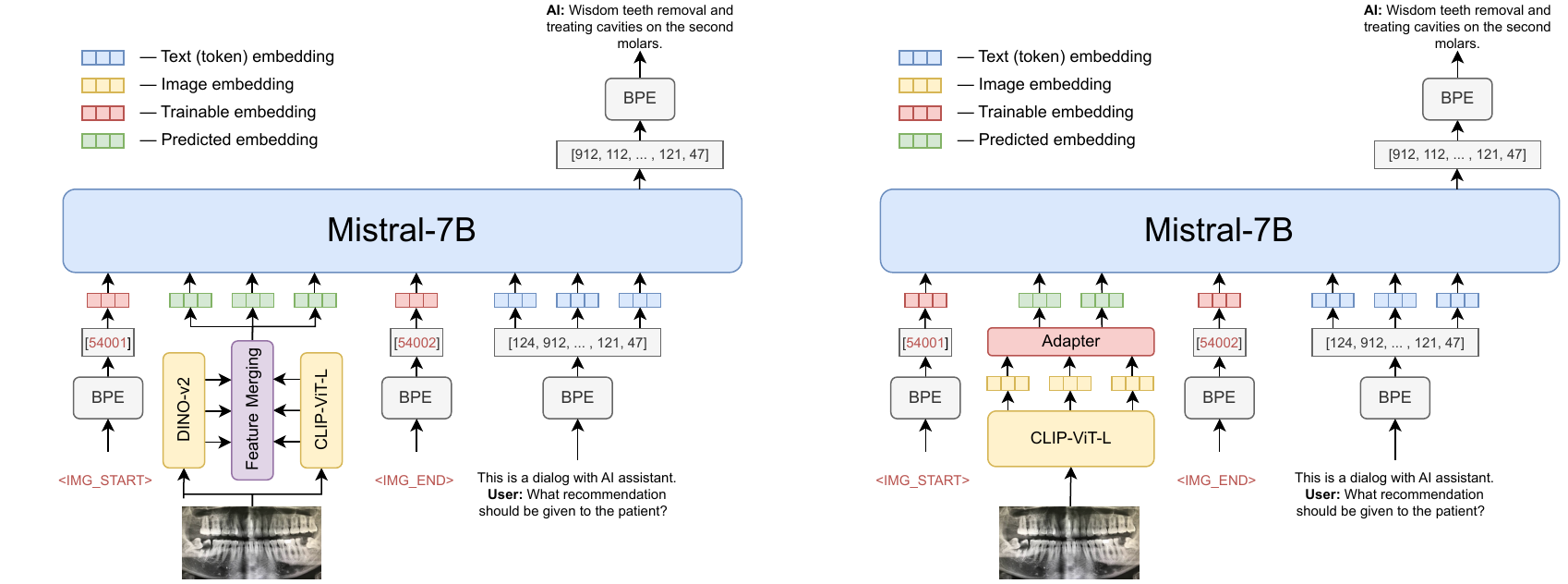}
    \caption{OmniFusion architecture with feature merging (left) and with single adapter (right): MLP or transformer layer.}
    \label{fig:OmniFusion}
\end{figure*}

The OmniFusion model integrates a pretrained LLM with special adapters for image embeddings, facilitating the fusion of the text and visual modalities. This approach has already proved its applicability in various VLM developments because the adapter approach is undemanding of computational resources in comparison with end-to-end training pipelines. Moreover, it does not need large interleaved text-image datasets to be trained, while for end-to-end approaches such data is extremely required. Overall, we have two major points that were investigated and covered further in this report: adapter technique selection and encoding visual data strategy determination. 

We used special tokens with trainable embeddings to signify the beginning and the end of a token sequence from the non-text modality. This approach enhances the model's ability to distinguish between textual and visual data streams. The visual encoder's output embeddings are processed either by a transformer adapter, consisting of a single transformer layer with four heads, or through a two-layer MLP.

Our experiments showed the most effective performance with a merging strategy that combines features of two encoders: CLIP-ViT-L and DINO-v2. The merging adapter is essentially a two-layer MLP, where the first layer is independent for each visual token, and the output layer is shared across all tokens. The merging strategy implies summing up the outputs after the first layer, optimizing the integration of textual and visual information (Figure~\ref{fig:OmniFusion}).

\subsection{Training pipeline}

When the overall architecture setup is fixed, we continue with overall training, which includes determination of the stages and the used datasets. The OmniFusion model undergoes a two-stage training process designed to harness its multimodal capabilities efficiently.

\paragraph{Stage 1: Pretraining.}

In the first stage, adapters and special tokens undergo pretraining on a vast dataset of image-text pairs. This phase aims to fine-tune the adapters for transforming visual embeddings and training special tokens to mark the transition between text and image data.

During this stage, we leverage image-text pairs tailored specifically for image captioning purposes. The image-captioning data we employ is sourced from ShareGPT4V-PT, which is supported by existing captioning datasets such as COCO and TextCaps. Additionally, we incorporate proprietary datasets containing document descriptions obtained through optical character recognition (OCR) in the pretraining set.

The instructions provided to the model during pretraining, include variations of prompts such as "Give a brief description of the image," "Describe the image in detail," and "Provide a short description of the image."

It's worth noting that we currently do not employ interleaved data during pretraining, which could potentially enhance the few-shot learning capabilities of multimodal models. However, we plan to explore this avenue in our future work.

We use the following sources of image-text captions: ShareGPT4V-PT (695K pairs), LAION-CC-SBU with BLIP captions (558K pairs) \cite{Liu2023ImprovedBW}. Overall, we utilize 1.2M image captions during the pretraining phase for the adapter alignment.

\paragraph{Stage 2: Fine-tuning.}

The second stage entails fine-tuning the multimodal model using a set of instructional dialogues. This process aims to enhance the model's ability to comprehend and respond to complex queries that require an integrated analysis of textual and visual information. Moreover, it is structured to facilitate multi-task fine-tuning in accordance with the natural instructions.

Recent research \cite{xu2024visionflan} has highlighted the potential pitfalls associated with the synthetic nature of data used for visual instruction tuning, as well as the lack of task diversity, which can lead to significant hallucinations in the output of multimodal models. To mitigate these challenges, we construct the supervised fine-tuning dataset by combining academic benchmarks with corresponding instructions and synthetic conversational data. This approach ensures a more diverse and robust training environment for the model. Table \ref{tab:sft-data} provides the datasets utilized in the fine-tuning procedure.

\begin{table}[h!]
\begin{center}
\begin{tabular}{ c | c | c}
 \hline
 Task & Dataset Source & Samples\\
 \hline
 Caption &  ShareGPT4V \cite{chen2023sharegpt4v} & 100K \\
 VQA & COCO \cite{chen2015microsoft}, SAM-9K \cite{kirillov2023segment}  & 20K, 9K\\
 WebQA & WebData \cite{cai2024internlm2} & 1.5K\\ 
 OCRQA & TextVQA \cite{singh2019towards}, OCRVQA \cite{mishraICDAR19} & 120K \\
 Conversation & LLaVA-v1.5-665K \cite{Liu2023ImprovedBW} & 665K \\
 DocVQA & Proprietary data (ru) & 20K \\
 \hline
 Text-only SFT & Proprietary data (ru), Alpaca (en) & 10K \\
 \hline
\end{tabular}
\end{center}
\caption{Distribution of the datasets for supervised fine-tuning.}
\label{tab:sft-data}
\end{table}

\paragraph{Training hyperparameters.}

During the training procedure, we utilize a standard approach with a learning rate of 1e-3 and a batch size of 256 for the pretraining phase. For the SFT phase, we adjust the learning rate to 2e-5 and the batch size to 128. We train the model using the bf16 precision, employing the AdamW optimizer with a weight decay set to 0.

The sequence length is initially set to 2048 for the base experiment, and we extend it to 4096 for experiments involving grid splitting. All experiments are conducted using 8 Nvidia A100-80Gb GPUs.

\begin{table}[ht]
\centering
\begin{tabular}{c|c|c}
\hline
\textbf{Parameter}            & \textbf{Pretraining Phase} & \textbf{SFT Phase} \\ \hline
Learning Rate                 & 1e-3                       & 2e-5               \\\hline
Batch Size                    & 256                        & 128                \\\hline
Precision                     & \multicolumn{2}{c}{bf16}                        \\\hline
Optimizer                     & \multicolumn{2}{c}{AdamW}                       \\\hline
Weight Decay                  & \multicolumn{2}{c}{0}                           \\\hline
Sequence Length               & \multicolumn{2}{c}{2048 (Base), 4096 (Grid Splitting)} \\\hline
Hardware                      & \multicolumn{2}{c}{8 Nvidia A100-80Gb GPUs}      \\ \hline
\end{tabular}
\caption{Training Procedure Parameters}
\label{tab:training_params}
\end{table}

\section{Experiments}

\subsection{Experimental setup}

\paragraph{Vision encoders.}

The influence of the image encoders is inevitable for multimodal systems. Basically, vision encoders are kept frozen via image-text alignment, so, its initial capabilities are crucial for further performance of LMMs. 

In our experiments, we evaluated several encoders varying in size, training data distribution, and image resolution. Overall, the visual encoders that we tested in the experiments, are listed below:

\begin{enumerate}
    \item CLIP ViT-bigG/14\footnote{https://github.com/mlfoundations/open\_clip}. A CLIP ViT-bigG/14 model was trained with the LAION-2B English subset of LAION-5B \cite{schuhmann2022laion5b} following the openclip architecture \cite{ilharco-gabriel-2021-5143773}.
    \item CLIP ViT-L/14 \cite{radford2021learning} from openai/clip-vit-large-patch14. This is the vision encoder that was retrieved from the CLIP model developed by researchers at OpenAI. The model is the standard version of vision encoders providing effective representation of the image features. The clip-L is used in such multimodal models as LLaVA-v1.5, LLaVA-Next, InternLM-XComposer2-VL \cite{dong2024internlmxcomposer2} and others.
    \item SigLIP/16-512\footnote{https://github.com/merveenoyan/siglip}. The SigLIP model was proposed in \cite{zhai2023sigmoid} by Google Research. SigLIP proposes to replace the loss function used in CLIP with a simple pairwise sigmoid loss. The model outperforms most of the existing models so far and provides state-of-the-art results in the classification tasks.
    \item InternViT-6B\footnote{https://github.com/OpenGVLab/InternVL-MMDetSeg}. InternViT-6B-448px-V1-2 \cite{chen2023internvl} is the largest open-sourced vision/vision-language foundation model to date, achieving 32 state-of-the-art performances on a wide range of tasks such as visual perception, cross-modal retrieval, multimodal dialogue, etc. The model was proposed by the OpenGVLab team.
    
    Pretrain Dataset: LAION-en, LAION-COCO, COYO, CC12M, CC3M, SBU, Wukong, LAION-multi, OCR data.
    
    The VLMs that utilize such encoders are as follows: InternVL-Chat-V1.2, InternVL-Chat-V1.2-Plus.

\end{enumerate}

Table \ref{tab:vision-enc} provides the overall description of the observed vision encoders that can be used for vision encoding in multimodal models.

\begin{table}[h!]
\begin{center}
\scriptsize
\begin{tabular}{ c | c | c | c | c | c | c }

 & patch resolution & fine-tuning resolution & num hidden layers & image tokens num & emb size & emb layer \\
 \hline
 CLIP \linebreak ViT-bigG/14 & 14x14 & 224x224 & 48 & 256 & 1664 & -2 \\
 \hline
 CLIP \linebreak VIT-large/14 & 14x14 & 336x336 & 24 & 576 & 1024 & -2 \\
 \hline
 SigLIP-base/16-512 & 16x16 & 512x512 & 12 & 1024 & 768 & -2 \\
 \hline
 InternViT-6B-448px-V1-2 & 14x14 & 448x448 & 45 & 1024 & 3200 & -1 \\
\end{tabular}
\end{center}
\caption{Vision encoders characteristics.}
\label{tab:vision-enc}
\end{table}

To evaluate the performance of different visual encoders on the multimodal model's visual understanding abilities, we trained the final model using the same data samples but with various vision backbones. We then assessed the model's quality on multimodal benchmarks to compare the effectiveness of different visual encoders. We present the evaluation results of OmniFusion trained with different vision backbones in Table \ref{tab:vision_enc_comp}.

\begin{table}[h!]
\begin{center}
\tiny
\begin{tabular}{ c | c| c | c | c | c | c | c | c | c | c }
  & Resolution & ScienceQA-full & ScienceQA-img & text-VQA & VQAv2 & VizWiz & POPE & MMBench & MMMU & MM-Vet \\
 \hline
CLIP ViT-bigG/14 & 336 & 68.71 & 66.68 & 47.58 & 73.28 & 49.92 & 82.80 & 61.73 & 30.60 & 30.10 \\
 \hline
CLIP VIT-large/14 & 336 & 71.21 & 68.17 & 57.20 & 78.42 & \textbf{54.11} & 84.50 & 65.72 & 30.90 & 33.60 \\
 \hline
SigLIP-base/16-512 & 512 & 71.92 & 68.62 & 57.05 & 78.88 & 51.84 & \textbf{86.5} & 66.80 & 31.60 & 33.70 \\
 \hline
InternViT-6B-448px-V1-2 & 448 & \textbf{73.38} & \textbf{69.71} & \textbf{61.57} & \textbf{80.08} & 52.78 & 84.47 & \textbf{67.10} & \textbf{35.40} & \textbf{35.10} \\
\end{tabular}
\end{center}
\caption{Comparison of OmniFusion performance with different vision backbones on multimodal benchmarks.}
\label{tab:vision_enc_comp}
\end{table}

Notably, the best performance is achieved by OmniFusion trained with the largest visual encoder, InternViT-6B-448px-V1-2. Despite the higher resolution for SigLIP, CLIP VIT-large/14 and SigLIP-base/16-512 achieve comparable results on most tasks. However, both of them significantly outperform CLIP ViT-bigG/14 with the smallest resolution.



 

\paragraph{Mix of image encoders.}

Rather than relying solely on a single image encoder, one approach to enhance image representation is to merge embeddings from multiple encoders. In this section, we considered various image encoders and adapters for mixing encoder features. We trained a language model and adapter using the features from multiple image encoders simultaneously. OpenAI CLIP-ViT-L/14 ($336\times336$) and ViT-DINO-v2 models were used as image encoders. Mixing of encoder features was performed in the adapter module. The following 4 different adapter module architectures were investigated.

The first architecture (baseline) corresponds to the encoder and adapter used in the previous experiments. It uses only one encoder (OpenAI CLIP-ViT-L/14) and an adapter with a transformer decoder layer. More specifically, the adapter unit has the following architecture. The model first uses the transformer encoder layer and then does a linear transformation to change the dimension of the embedding vectors.

The second architecture (embedding concatenation) uses two encoders, first, the vectors of the feature sequences from the penultimate layers of encoders are independently linearly mapped. Two different linear transformations are used for the features from the first and the second encoders. Afterward, the embedding sequences are then concatenated and fed to the encoder layer of the transformer and mapped linearly (to change the dimension of the embeddings) as in the baseline.

The third architecture, inspired by COMM \cite{jiang2024clip}, also uses 2 encoders. An important difference from the previous version is the use of the features from all layers of image encoders.
First, layer normalization is applied to the feature from all layers of each encoder, then linear transformation is applied (different mappings are used for the features from the different layers), after which, a linear combination of the features from different layers is calculated (the coefficients of the linear combination are trainable parameters), the resulting features for two encoders are summed (sequences of embeddings from two encoders have different lengths, so the shorter sequence is concatenated with zero vectors before the addition). Finally, the GELU activation function and the final linear transformation are applied.

The last architecture, the fourth one, is similar to the third, but after aggregating features from different layers of encoders (counting linear combinations of features from different layers of encoders), it uses a different method of mixing features from two encoders. GELU is applied to the embedding sequences from two encoders, then the embedding sequences are concatenated and fed to the multi-head attention layer as a query, and a trained parameter matrix is used for the key and value. In the end, the GELU transform and linear mapping are applied. This approach is interesting because by changing the size of the trained parameter matrix, it is possible to change the number of vectors in the output embedding sequence. We used a matrix of 576 rows so that the length of the visual embedding sequence is the same as in the baseline.

Table \ref{tab:vision-enc-mix} presents the results of the tests on various benchmark models that use different methods of mixing features from two visual encoders.

\begin{table}[h!]
\begin{center}
\scriptsize
\begin{tabular}{ c | c | c | c | c | c | c | c | c | c }

  & ScienceQA-full & ScienceQA-img & Text-VQA & VQAv2 & VizWiz & POPE & MMBench & MMMU & MM-Vet \\
 \hline
 Baseline (one encoder) & 71.21 & 68.17 & 57.20 & 78.42 & 54.11 & 84.50 & 65.72 & 32.30 & 33.60 \\
 \hline
 Embeddings concatenation & \textbf{72.25} & \textbf{68.62} & 54.02 & 75.95 & 51.33 & 85.40 & 67.87 & 34.10 & 30.60 \\
 \hline
 Projection \& summation & 71.09 & 67.18 & \textbf{60.87} & \textbf{78.77} & \textbf{57.09} & \textbf{85.73} & \textbf{69.24} & \textbf{36.50} & \textbf{34.60} \\
 \hline
 Projection \& multi-head attention & 72.01 & 68.17 & 47.75 & 70.84 & 50.82 & 84.93 & 57.47 & 29.40 & 21.90 \\

\end{tabular}
\end{center}
\caption{Comparison of OmniFusion performance with different CLIP VIT-L/14 and DinoV2 encoders mix.}
\label{tab:vision-enc-mix}
\end{table}

The best method turned out to be one that uses features from all the layers of encoders and sums them to mix the features from two encoders. We also noticed that the last approach with a multi-headed attention layer turned out to be the worst in terms of the inference speed.

In the initial experiments, we also compared two adapter techniques: linear projection, implemented as a two-layer MLP with GeLU activation, and a simple transformer-based Encoder block with one layer and four heads. We found that the best results were obtained with the simple linear projection. While the transformer-based adapter performs comparably to MLP on some benchmarks (such as VizWiz or ScienceQA), we observed a drop of up to 4 points on OCR-based tasks.

We hypothesize that this difference can be explained by the fact that originally, Vision Transformer (VIT) models trained in the CLIP pipeline are naturally aligned with text, making linear transformation suitable for this task. Additionally, we speculate that larger trainable adapters require more data for successful alignment.

\paragraph{Scaling Images to HD.}

Following the methodology of LLaVA-NeXT \cite{liu2024llavanext}, we investigate whether image resolution plays a significant role in capturing fine-grained pixel scale information.

Our approach involved resizing images adaptively to the input size and dividing them into non-overlapping sections based on the resolution of the image encoder. This strategy ensured that we retained all the specific details from the original images, thus enhancing the model's performance on tasks that rely on such details.

To encode the image, we employ CLIP-ViT-L/14 ($336\times336$), treating each patch as a separate entity. Initially, we resized and padded the image to the most suitable dimensions. Then, we divided it into a grid of $336\times336$ patches, encoding each patch individually. As has been observed in previous research, encoding the entire image into a sequence of image tokens is crucial for preserving the overall context of the scene, so, we also kept this image representation in our final experiment.

To assess the model's capability to encode images with higher resolution. we trained three versions of the same model. The baseline model is OmniFusion-7B with the frozen OpenAI Clip-L vision encoder. The second model follows the same architecture, but during fine-tuning, the image is split into patches and encoded by the Clip-L encoder. Finally, we augmented the training data with 20K proprietary samples from the DocVQA domain to evaluate whether this additional dataset, combined with image splitting, enhances the model's performance in the document domain.

Table \ref{tab:grid} presents the results of OmniFusion trained with the strongest vision backbone in standard resolution ($448\times448$), and in high resolution with CLIP-ViT-L/14 ($336\times336$), and grid split on standard multimodal benchmarks, while Table \ref{tab:info} shows the results for specific benchmarks related to document and infographics analysis.

\begin{table}[h!]
\begin{center}
\scriptsize
\begin{tabular}{c | c | c | c | c | c | c | c | c | c | c }
  & \makecell{ScienceQA\\full} & \makecell{ScienceQA\\img} & \makecell{text-VQA \\w OCR} & \makecell{Text-VQA \\w/o OCR} & VQAv2 & POPE & MMBench & MMMU & MM-Vet & GQA \\
 \hline
\makecell{OmniFusion \\ InternViT-6B-448px-V1-2} & \textbf{73.38} & \textbf{69.71} & 61.57 & 56.48 & 80.08 & 84.47 & 67.10 & 35.40 & 35.10 & 63.44 \\
 \hline
 \makecell{OmniFusion \\ grid split} & 71.23 & 67.97 & 63.31 & 60.06 & 80.86 & 86.72 & \textbf{69.00} & 34.10 & 36.60 & 64.18 \\
 \hline
\makecell{OmniFusion \\ grid split + ruDocVQA} & 73.21 & 69.16 & \textbf{65.53} & \textbf{63.50} & \textbf{80.94} & \textbf{87.21} & 67.61 & \textbf{36.60} & \textbf{39.40} & \textbf{64.57} \\

\end{tabular}
\end{center}
\caption{Comparison of OmniFusion performance w and w/o grid splitting of the input image.}
\label{tab:grid}
\end{table}

\begin{table}[h!]
\begin{center}
\scriptsize
\begin{tabular}{c | c | c | c | c }
  & InfoVQA & ChartQA & DocVQA & multiDocVQA \\
 \hline
\makecell{OmniFusion \\ InternViT-6B-448px-V1-2} & 29.24 & 19.92 & 38.44 & 22.5/11.43 \\
 \hline
 \makecell{OmniFusion \\ grid split} & 30.2 & 20.2 & 46.76 & 27.09/14.5 \\
 \hline
\makecell{OmniFusion \\ grid split + rudocVQA} & \textbf{36.99} & \textbf{28.96} & \textbf{62.82} & \textbf{36.38/20.55} \\

\end{tabular}
\end{center}
\caption{Comparison of OmniFusion w and w/o grid splitting for document and infographics analysis.}
\label{tab:info}
\end{table}

Based on the observed results, we make the following observations:

\begin{enumerate}
    \item First, high image resolution is particularly relevant in domains such as OCR. Additionally, the evaluation of hallucination (POPE) is also influenced by image resolution.

    \item Another interesting finding concerns document-based benchmarks. Adding 20K samples of proprietary document-level data in Russian significantly boosted the model's performance on document analysis (refer to Table \ref{tab:info}, line 3).
\end{enumerate}

\paragraph{Tuning with synthesized TeX formulas.}

In this section, we examined SFT for the task of generating LaTeX code from a formula image. The datasets used for training included CROHME \cite{CROHME} (containing pairs of handwritten formulas and LaTeX code) and the im2latex dataset (containing pairs of formulas rendered in different fonts and their corresponding LaTeX code). Both of these preprocessed datasets were obtained from the open-source repository LaTeX-OCR\footnote{https://github.com/lukas-blecher/LaTeX-OCR}. Overall, the character sequence length in the training dataset ranges from 1 to 4296 symbols.

As image encoders we used CLIP VIT-large/14 and Donut \cite{kim2022ocr} vision encoder. The last was fine-tuned for the text and LaTeX formula recognition in the open-source repository texify\footnote{https://github.com/VikParuchuri/texify} and provided a good latent representation for this specific case of math equation recognition.

SFT stage consisted of two steps:

\begin{enumerate}
    \item Freezing the image encoder and LLM, fine-tuning the adapter and the special tokens.
    \item Freezing the image encoder, fine-tuning the adapter, and the special tokens, and fine-tuning LLM using LoRA.
\end{enumerate}

As a result of the study, when using CLIP VIT-large/14 as the image encoder, the model performs poorly in long formula recognition but is capable of perceiving the handwritten ones, whereas the Texify vision encoder shows the opposite behavior. The comparison of the two encoders for TeX formula recognition is presented in Table \ref{tab:tex}.

The normalized edit distance (NED) was utilized as a metric to evaluate the quality of LaTeX code generation across 30,600 samples from the im2latex test dataset, encompassing formulas ranging in length from 4 to 2226 symbols.

\begin{table}[h!]
\small
\begin{center}
\begin{tabular}{ l | c }
    Image Encoders & NED $\downarrow$ \\
 \hline
    CLIP VIT-large/14 & 0.32 \\
 \hline
    Texify vision encoder & \textbf{0.19} \\
\end{tabular}
\end{center}
\caption{Comparison of two image encoders for OmniFusion for TeX formulas recognition.}
\label{tab:tex}
\end{table}

\begin{figure}[h!]

    \centering
    \includegraphics[scale=0.6]{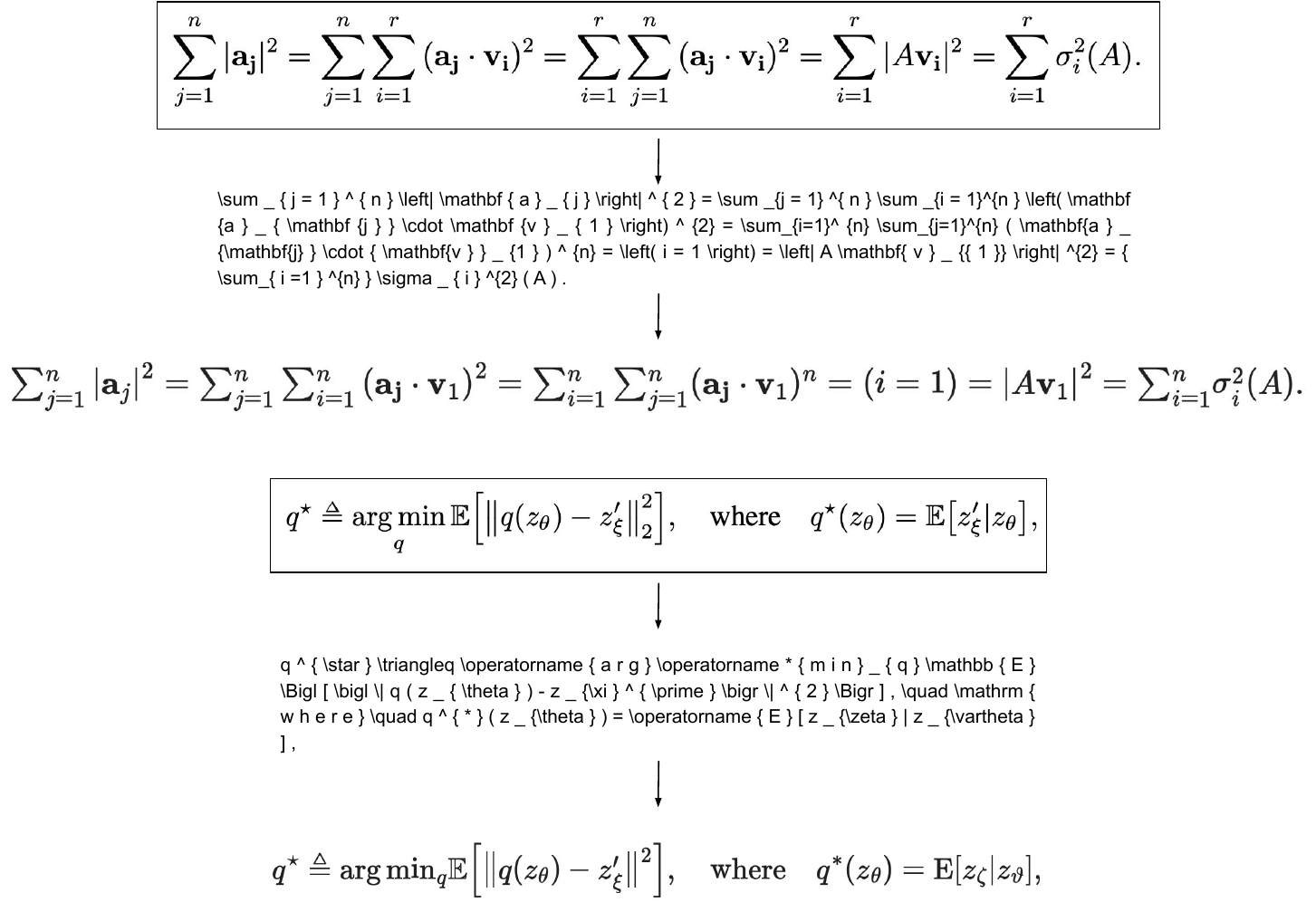}
    \caption{An example of LaTeX formula understanding by OmniFusion fine-tuned with the Texify vision encoder is depicted below. The upper image displays the input image, while the lower image showcases the compilation of LaTeX code generated by the model.}
    \label{fig:text-fig}
\end{figure}

Figure \ref{fig:text-fig} provides examples of LaTeX formula understanding by OmniFusion with the Texify vision encoder.


\subsection{Evaluation on benchmarks}




In this section, we present the evaluation results of the introduced model. Table \ref{tab:model-comp} presents a comparison of existing leading approaches to training multimodal models with our best options of OmniFusion. We provide a comparison of the models with standard and high resolution across several benchmarks in zero-shot settings. For the evaluation of OmniFusion and LLaVA, we utilize the recent open-source library lmms-eval \cite{lmms-eval2024}.

\begin{table}[h]
\begin{center}
\caption{Comparison with recent multimodal models on zero-shot benchmarks. The best results across model subgroups are in bold; the best performance for models with LLMs of the same size is underlined. As we lack information on how TextVQA is assessed, for some models we include additional evaluation with supported OCR tokens in brackets. We do not provide information on whether the train set of the benchmark was included.}
\label{tab:model-comp}
\tiny
\begin{tabular}{c | c | c | c | c | c | c | c | c | c | c | c}

 Model & LLM & Encoder & Res. & \makecell{ScienceQA \\ full} & \makecell{ScienceQA \\ img} & Text-VQA & VQAv2 & POPE & GQA & MMMU &  MM-Vet \\
  \hline
 \multicolumn{12}{c}{\textit{Standard resolution}} \\
 \hline
  LLaVA-v1.5 13B \cite{Liu2023ImprovedBW} & Vicuna-13B & CLIP VIT-large/14 & 336 & \textbf{74.96} & \textbf{72.88} & \makecell{48.73 \\ (60.75)} & 79.52 & 85.92 & 63.24 & 34.80 & 36.70 \\
  \hline
 LLaVA-v1.5 7B \cite{Liu2023ImprovedBW} & Vicuna-7B & CLIP VIT-large/14 & 336 & 70.41 & 70.43 & \makecell{46.07 \\ (58.20)} & 78.50 & 85.87 & 61.97 & 35.30 & 30.55	\\
 \hline
 Qwen-VL-Chat \cite{Bai2023QwenVLAV} & Qwen-7B & CLIP VIT-bigG/14 & 448 & - & 68.2 & 63.80 & - & \underline{\textbf{88.10}} & 57.50 & \underline{\textbf{37.00}} & \underline{\textbf{47.30}} \\
  \hline
 Mini-Gemini \cite{li2024minigemini} & Vicuna-7B & \makecell{CLIP VIT-large/14 \\ ConvNeXt-L} & 336 & - & - & \underline{\textbf{65.20}} & - & \underline{\textbf{88.10}} & - & 36.10 & 40.80 \\
 \hline
 \textbf{OmniFusion} & GigaChat-7B & CLIP VIT-large/14 & 336 & 71.21 & 68.17 & \makecell{51.87 \\ (57.20)} & 78.42 & 84.50 & \textbf{65.72} & 32.30 & 33.60 \\
 \hline
 \textbf{OmniFusion} & GigaChat-7B & \makecell{InternViT-6B-448px \\ V1-2} & 448 & \underline{74.13} & \underline{71.29} & \makecell{56.48 \\ (61.57)} & \underline{\textbf{80.08}} & 84.14 & 63.44 & 35.40 & 35.10 \\
 \hline
 \multicolumn{12}{c}{\textit{Encoder Merge}} \\
 \hline
  \textbf{OmniFusion} & GigaChat-7B & \makecell{CLIP VIT-large/14 \\+DinoV2} & 336 & 71.09 & 67.18 & \makecell{49.75 \\ (60.87)} & 78.77 & 85.73 & \underline{64.59} & 36.50 & 34.60 \\
   \hline
    \textbf{OmniFusion} & Mistral-7B & \makecell{CLIP VIT-large/14 \\+ DinoV2} & 336 & 70.30 & 67.55 & \makecell{48.93 \\ (59.15)} & - & 82.18 & 64.00 & 35.90 & 33.10 \\
   \hline
   \textbf{OmniFusion} & Vicuna-7B & \makecell{CLIP VIT-large/14 \\+ DinoV2} & 336 & 71.26 & 69.96 & \makecell{49.58 \\ (61.04)} & - & 85.81 & 62.60 & \underline{36.90} & 33.60 \\
 \hline
 \multicolumn{12}{c}{\textit{HR resolution}} \\
 \hline
  LLaVA-Next 7B & Vicuna-7B & CLIP VIT-large/14 & 672 & \underline{73.21} & \underline{70.15} & \underline{64.85} & 80.06 & 86.40 & 64.23 & 35.10 & \underline{\textbf{44.08}} \\
 \hline
 LLaVA-Next 13B & Vicuna-13B & CLIP VIT-large/14 & 672 & \textbf{75.85} & \textbf{73.57} & \textbf{66.92} & 80.92 & 86.26 & \textbf{65.36} & 35.90 & \textbf{49.12} \\
 \hline
 DeepSeek-VL \cite{lu2024deepseekvl} & DeepSeek-7B & \makecell{SigLIP-L \\ SAM-B} & \makecell{384 \\ 1024} & - & - & - & - & 88.10 & - & \underline{36.6} & 41.50 \\
 \hline
 Mini-Gemini-HD \cite{li2024minigemini} & Vicuna-7B & \makecell{CLIP VIT-large/14 \\ ConvNeXt-L} & 672 & - & - & 68.40 & - & - & - & \underline{\textbf{36.8}} & 41.30 \\
  \hline
 \makecell{OmniFusion \\ + grid-split} & GigaChat-7B & CLIP VIT-large/14 & 672 & \underline{73.21} & 69.16 & \makecell{63.5 \\ (65.53)} & \textbf{\underline{80.94}} & \underline{\textbf{87.21}} & \underline{64.57} & \underline{36.60} & 39.40 \\
 \hline
 \multicolumn{12}{c}{\textit{Proprietary models}} \\
 \hline
  GPT-4V & Unk & Unk & Unk & - & - & 78.00 & - & - & - & \textbf{56.80} & 49.90 \\
  \hline
  Gemini Pro & Unk & Unk & Unk & - & - & 74.60 & - & - & - & 47.90 & \textbf{64.30} \\
  \hline
  Qwen-VL-Plus & Unk & Unk & Unk & - & - & 78.90 & - & - & - & 45.20 & 55.70 \\
  \hline
  Qwen-VL-Max & Unk & Unk & Unk & - & - & \textbf{79.50} & - & - & - & 51.40 & 61.80 \\
\end{tabular}
\end{center}
\end{table}

\paragraph{Main results.} The main results are as follows: at standard resolution, OmniFusion using the largest vision encoder InternViT-6B-448px-V1-2 achieves the best scores on most of the benchmarks, being on an equal basis with models based on larger LLMs (such as Vicuna-13B \cite{zheng2023judging}).

Moreover, in the subgroup of models that operate on images with the same resolution, the proposed approach of encoder mixing significantly improves results in Text-VQA and GQA datasets, adding one point, on average, to the MMMU benchmark as well.

For high resolution, the version of OmniFusion with grid splitting provides comparative results with state-of-the-art open-source multimodal models.

\section{Related Work}

Taking advantage of LMMs powerful capability to follow instructions, active work is underway to extend their ability to perceive other types of data, particularly visual data. Early work in the field, including FROMAGe \cite{Koh2023GroundingLM}, LLaVA \cite{Liu2023VisualIT}, MiniGPT4 \cite{Zhu2023MiniGPT4EV}, and BLIP-2 \cite{Li2023BLIP2BL}, encoded visual information into one or more visual tokens by training only a lightweight projection that efficiently matched images to the language model. This approach has enabled tasks such as image captioning and visual question answering to be partly solved. Several subsequent studies have focused on enhancing and adjusting the projection architecture \cite{Dai2023InstructBLIPTG,Alayrac2022FlamingoAV,Jian2023BootstrappingVL,Lu2023LyricsBF}, and determining the need for pretraining LLM on image text data \cite{Zhang2023LLaMAAdapterEF,Gao2023LLaMAAdapterVP,Bai2023QwenVLAV,Zhou2023InfMLLMAU}. Another promising approach was introduced in \cite{wang2024cosmo,peng2023kosmos2}, the authors showed that being pre-trained from scratch on large amount of interleaved data multimodal models obtain external abilities to understand long context and support few-shot learning. However, since training these models requires a lot of computational resources there's greater interest in lighter-weight approaches. These involve partially freezing specific parts of the backbones within the multimodal pipeline \cite{chen2023pali,Liu2023VisualIT,Bai2023QwenVLAV}.

Significant improvement in visual-textual perceptual quality was achieved by curating more synthetic fine-tuning data, as introduced in LLaVA-1.5 \cite{Liu2023ImprovedBW}. This model has a relatively simple architecture and uses only publicly available datasets, highlighting the importance of data and multimodal task diversity for VLM training. A number of subsequent papers have improved this approach by enhancing the visual grounding capability in LLaVA-G \cite{Zhang2023LLaVAGroundingGV}, adapting to specific domains as LLaVA-Med \cite{Li2023LLaVAMedTA}; instructively learning to use tools and invoking other task-specific models in LLaVA-PLUS \cite{Liu2023LLaVAPlusLT}, improving efficiency through MoE-based sparse modification in MoE-LLaVA \cite{Lin2024MoELLaVAMO}, and reducing the underlying language model to Phi-2.7B in the LLaVA-Phi paper \cite{Zhu2024LLaVAPhiEM}. The Video-LLaVA work \cite{Lin2023VideoLLaVALU} continues to expand the horizons of multimodality to the video comprehension.

At the same time, state-of-the-art results have been achieved through careful investigation of the most beneficial process of visual-textual pre-learning and the effective incorporation of the visual modality into VLLM. The VILA paper \cite{Lin2023VILAOP} argues that unfreezing the LLM allows for stronger in-context learning and multi-image reasoning capabilities. In addition, these results also further support the conclusions of the Monkey paper that the image resolution is more important than the number of visual tokens. Such an approach to embedding visual modality may help to provide greater scalability and better quality on fine-grained visual-language tasks. COMM \cite{jiang2024clip} revised the effectiveness of already existing visual models in MLLM and proposed a simple yet effective multi-level features merging strategy for different image encoders to enhance visual capabilities. The follow-up VistaLLM work \cite{Pramanick2023JackOA} proposed an instruction-guided image tokenizer that filters global embeddings using task descriptions to extract compressed and refined features from images into a unified general-purpose model.

Despite these advancements, there are still open challenges in extracting better visual features from images, indicating the need for further development of multimodal architectures in this area. Our work takes mutual advantages of both previously mentioned research directions, i.e., using the LLaVA-based VLM architecture as a baseline and effectively combining multiple visual encoders into a single visual branch providing a good opportunity to enhance latent visual representation for general and specific domains.

\section{Conclusion}

In conclusion, our OmniFusion model represents significant progress in the field of multimodal learning, integrating textual and visual data within a single framework. By evaluating various architecture design principles, and image encoding methods, and integrating multiple vision encoders, we have demonstrated the model's superior performance across a wide array of visual-language benchmarks. Our findings underscore the effectiveness of OmniFusion in handling complex VQA tasks, outperforming existing solutions, and providing detailed, domain-specific responses. With the release of our open-source Mistral-based OmniFusion model, including weights and scripts for training and inference, we aim to contribute to the broader AI research community, facilitating further developments in multimodal AI systems.

Further development of our research will shed light on the new approaches of efficient image embedding extraction including not only encoders research but also some hierarchical image representation methods. In terms of overall architecture design, we plan to go deeper into long image context processing, especially in video modality integration. At the same time, providing fast and reliable image output generation using our latent diffusion model Kandinsky \cite{razzhigaev-etal-2023-kandinsky} is also one of the major research tasks, which will lead to the new OmniFusion features like local image editing by text prompt, image generation during multimodal dialog context, etc.

\bibliographystyle{unsrt}
\bibliography{references}

\newpage
\phantomsection
\section*{A. Acknowledgements}

The authors would also like to extend their deepest appreciation to SberAI and Sber Devices teams for providing help with multimodal data engineering, GigaChat large language models checkpoints, and invaluable advice in OmniFusion research (G. Novikov, S. Markov, F. Minkin, et al.).

\section*{B. OmniFusion Examples}

Below there are few examples of OmniFusion generations in English and Russian.

\centering
\includegraphics[scale=0.6]{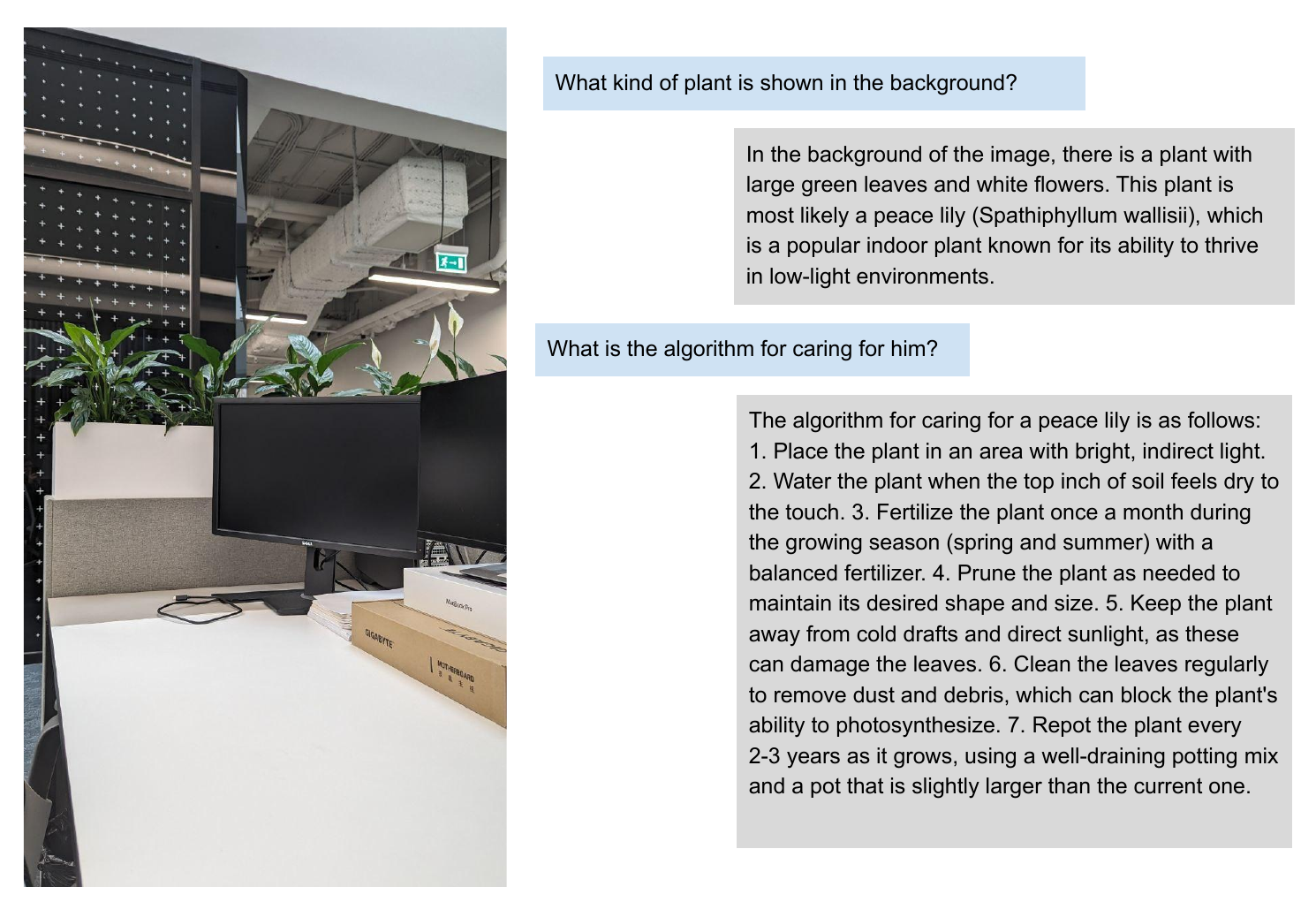}

\vspace{1cm}

\centering
\includegraphics[scale=0.6]{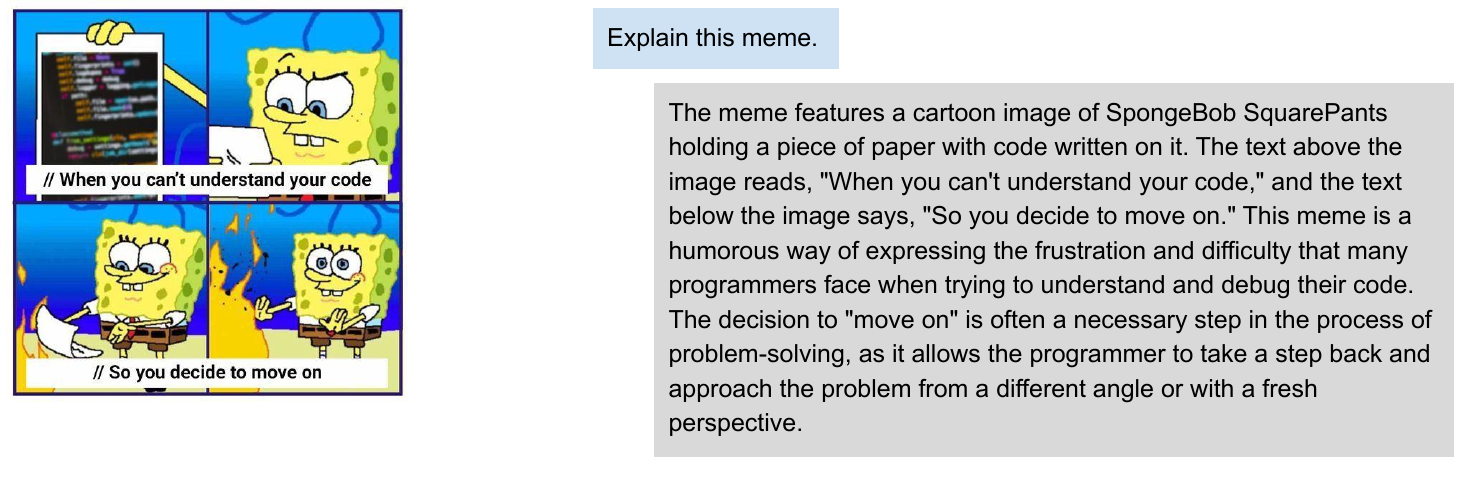}

\vspace{1cm}

\centering
\includegraphics[scale=0.6]{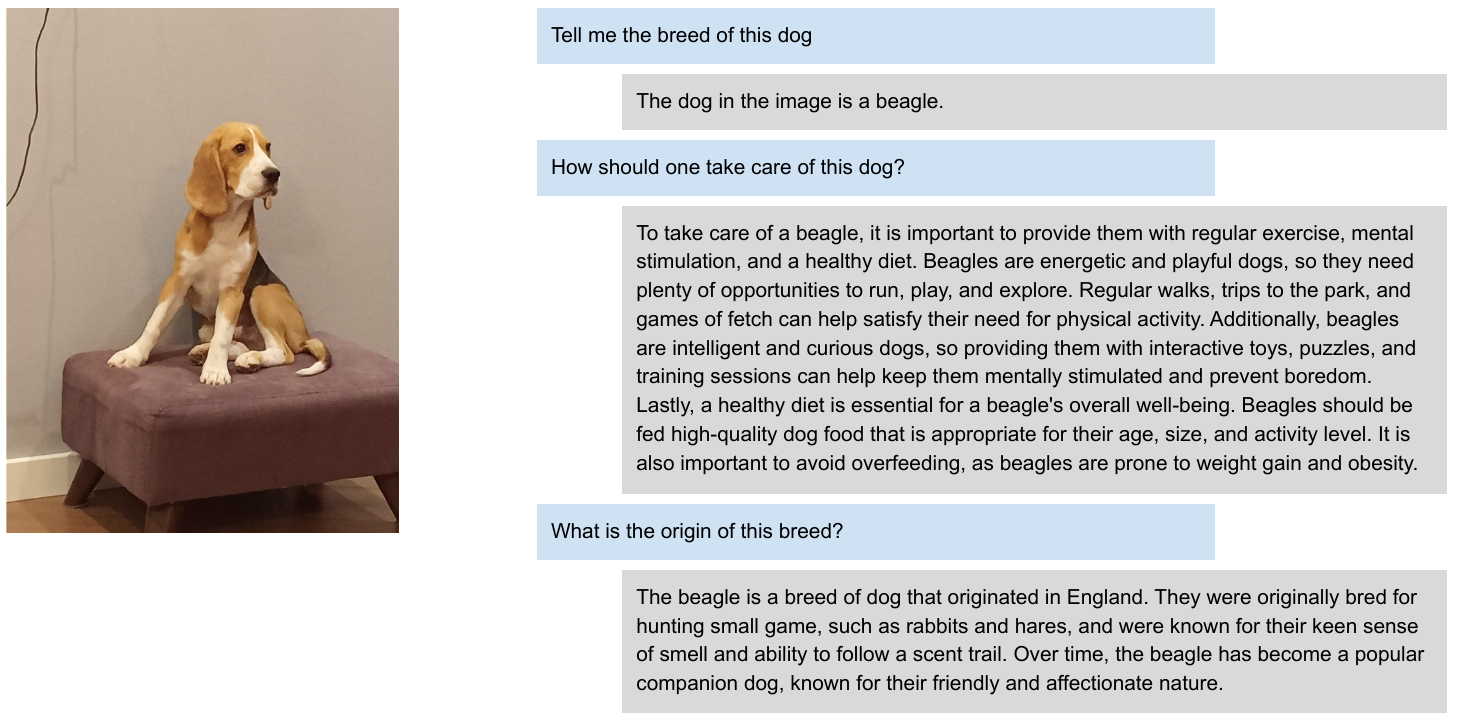}

\vspace{1cm}

\centering
\includegraphics[scale=0.6]{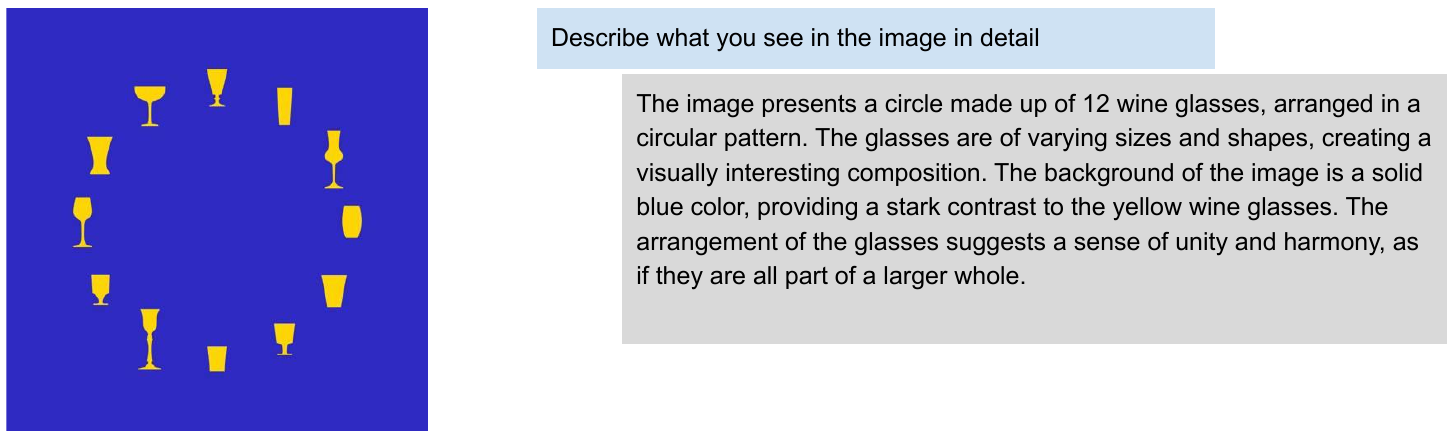}

\vspace{1cm}

\centering
\includegraphics[scale=0.6]{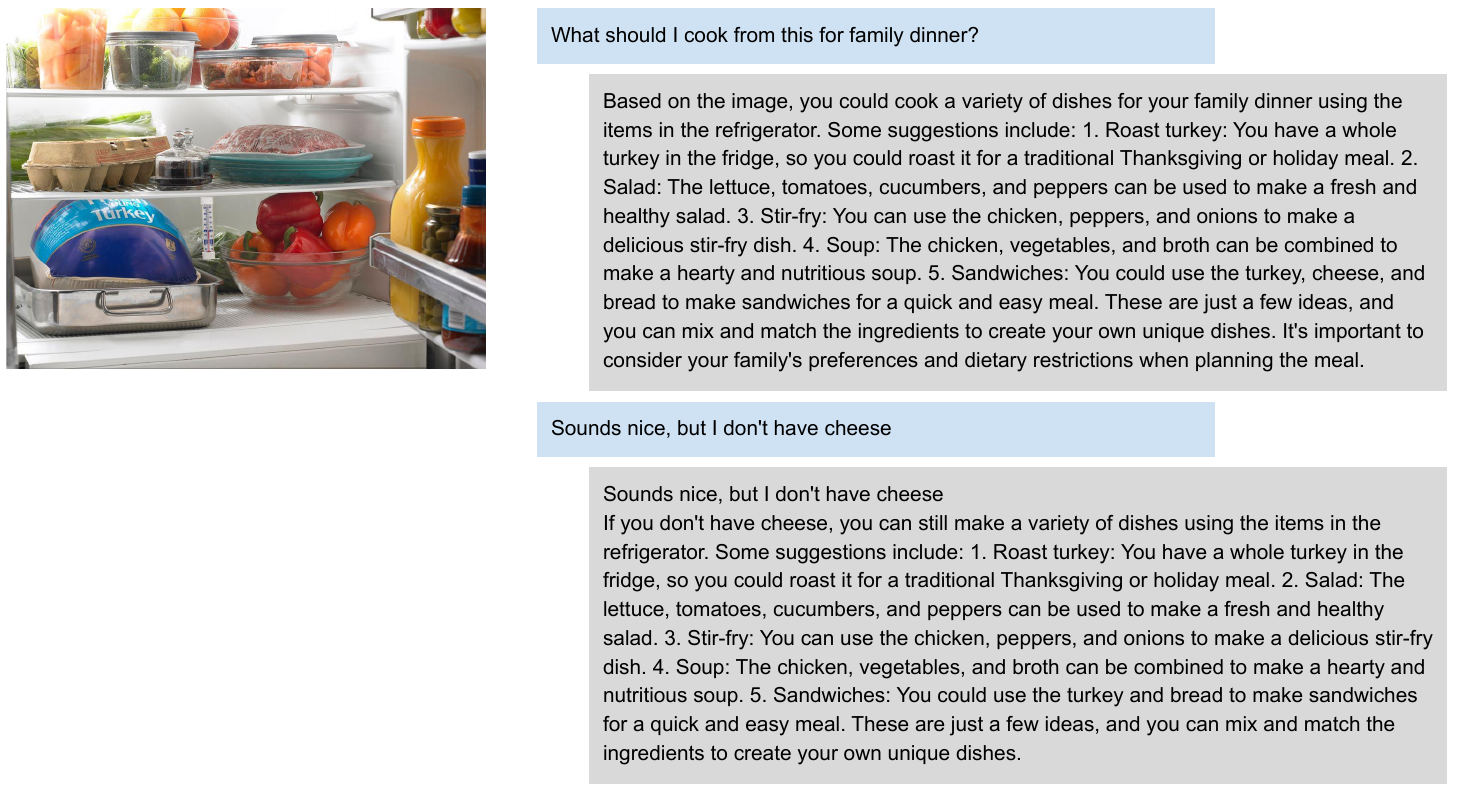}

\vspace{1cm}

\centering
\includegraphics[scale=0.6]{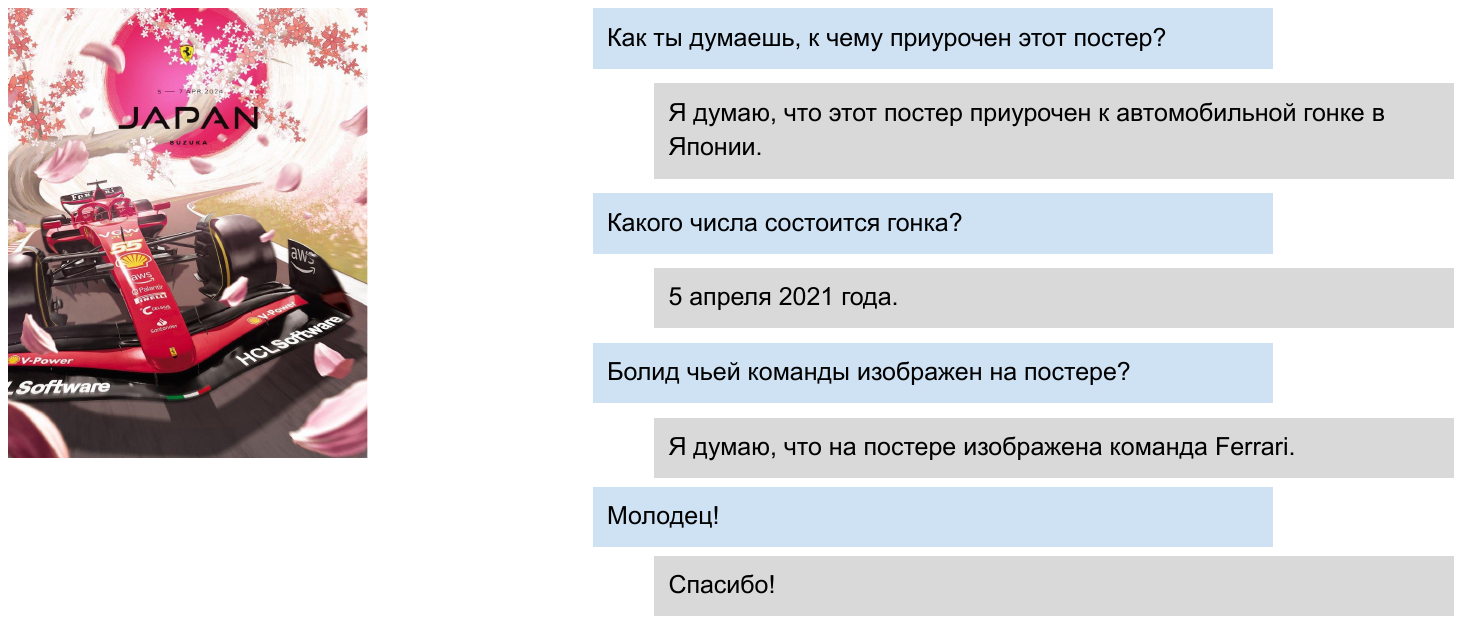}

\vspace{1cm}

\centering
\includegraphics[scale=0.6]{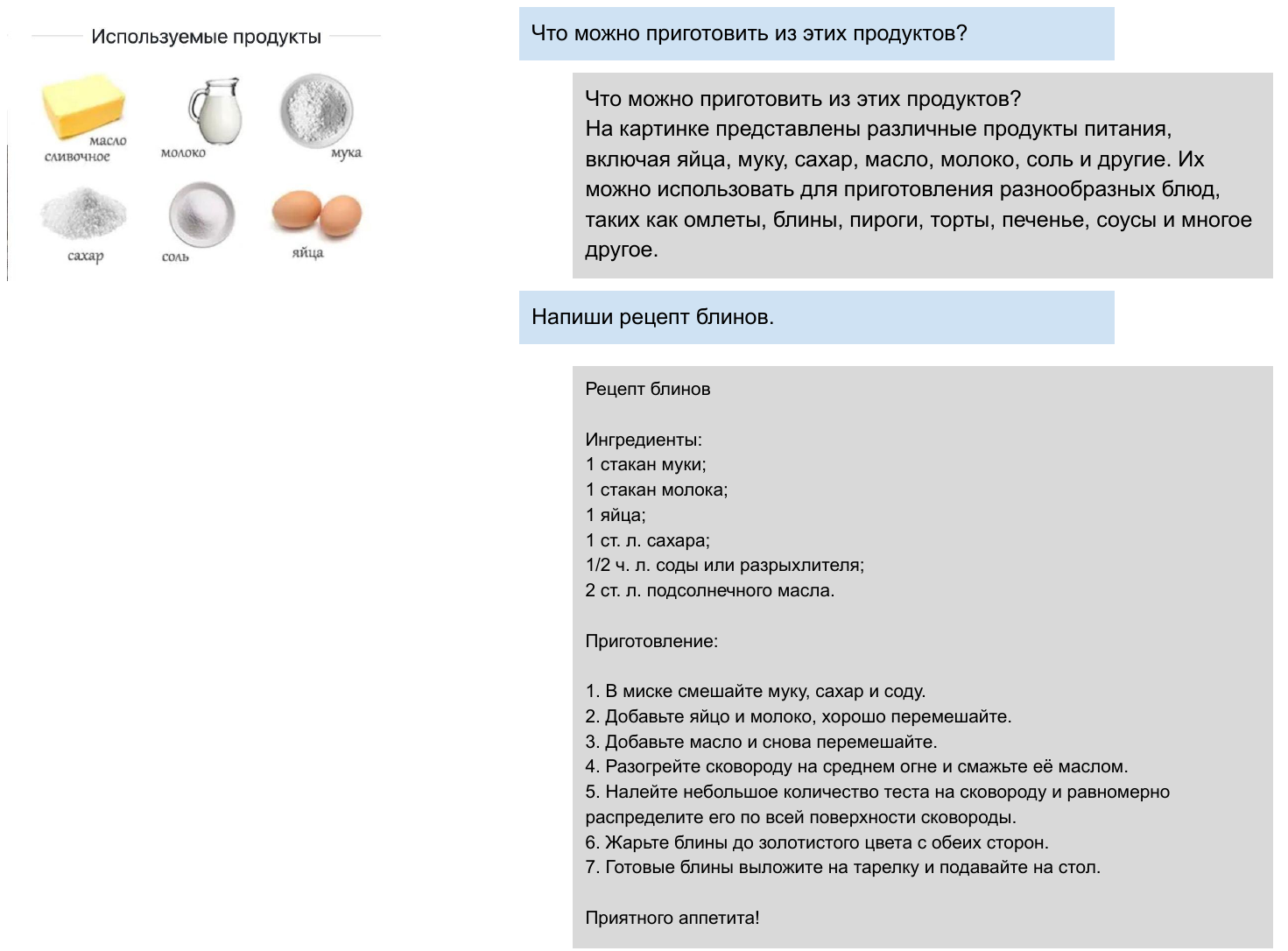}

\vspace{1cm}

\centering
\includegraphics[scale=0.6]{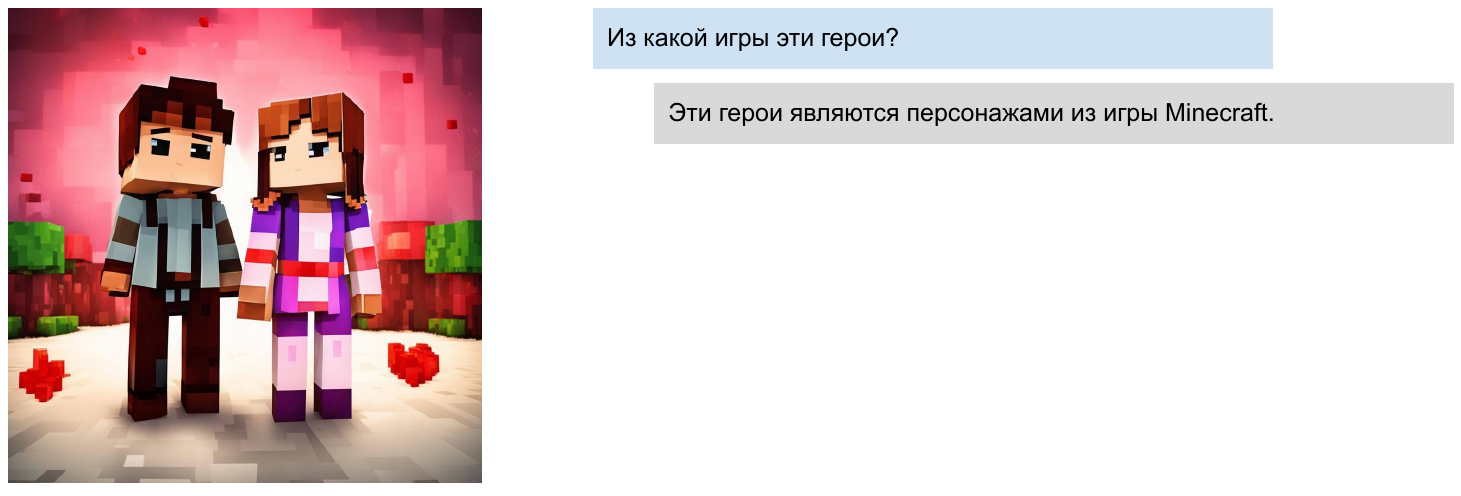}

\vspace{1cm}

\centering
\includegraphics[scale=0.6]{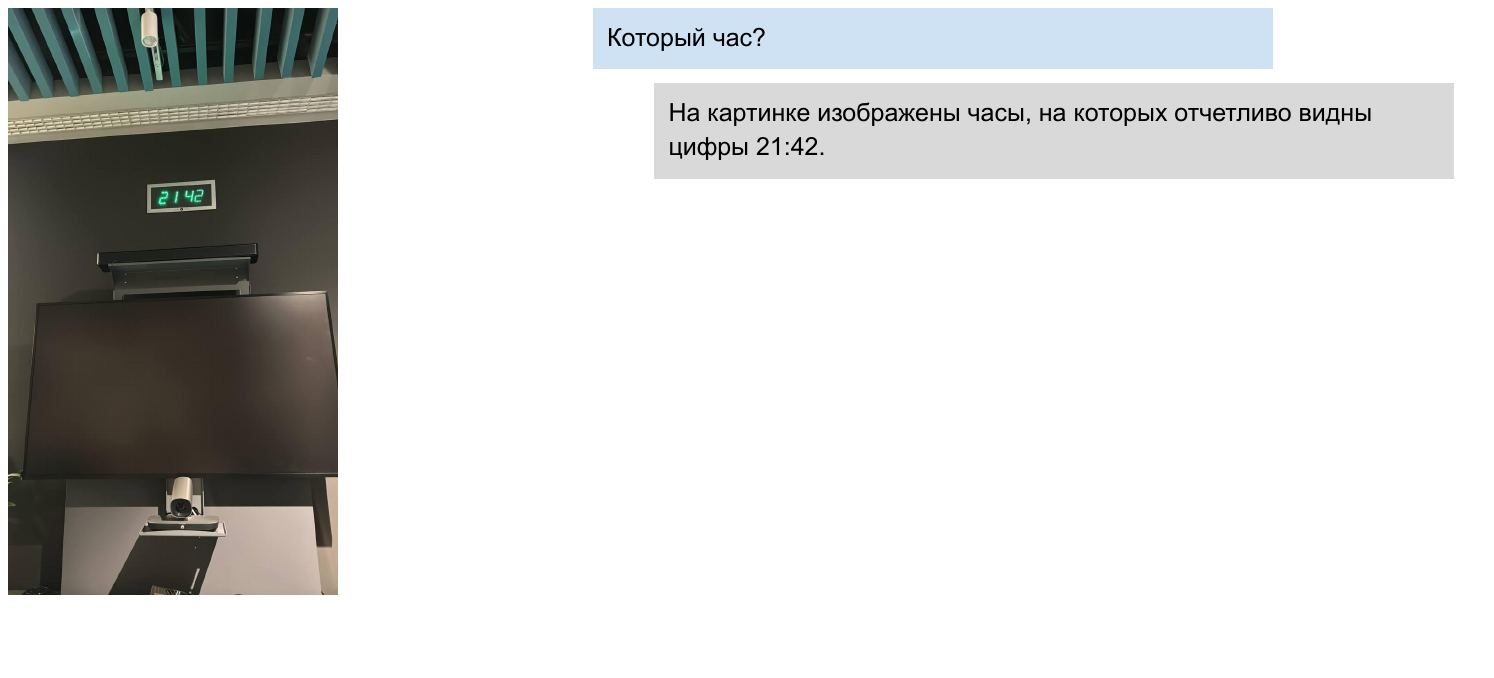}

\vspace{1cm}

\centering
\includegraphics[scale=0.6]{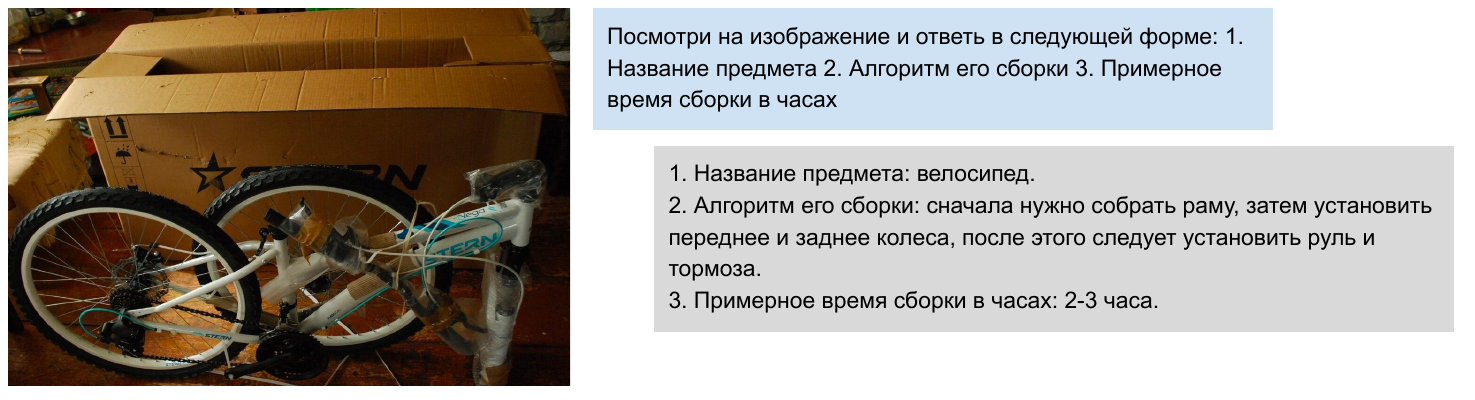}

\end{document}